# Causal Prototype-inspired Contrast Adaptation for Unsupervised Domain Adaptive Semantic Segmentation of High-resolution Remote Sensing Imagery

Jingru Zhu, Ya Guo, Geng Sun, Liang Hong and Jie Chen, *Member, IEEE*

*Abstract*—Semantic segmentation of high-resolution remote sensing imagery (HRSI) suffers from the domain shift, resulting in poor performance of the model in another unseen domain. Unsupervised domain adaptive (UDA) semantic segmentation aims to adapt the semantic segmentation model trained on the labeled source domain to an unlabeled target domain. However, the existing UDA semantic segmentation models tend to align pixels or features based on statistical information related to labels in source and target domain data, and make predictions accordingly, which leads to uncertainty and fragility of prediction results. In this paper, we propose a causal prototype-inspired contrast adaptation (CPCA) method to explore the invariant causal mechanisms between different HRSIs domains and their semantic labels. It firstly disentangles causal features and bias features from the source and target domain images through a causal feature disentanglement module. Then, a causal prototypical contrast module is used to learn domain invariant causal features. To further de-correlate causal and bias features, a causal intervention module is introduced to intervene on the bias features to generate counterfactual unbiased samples. By forcing the causal features to meet the principles of separability, invariance and intervention, CPCA can simulate the causal factors of source and target domains, and make decisions on the target domain based on the causal features, which can observe improved generalization ability. Extensive experiments under three cross-domain tasks indicate that CPCA is remarkably superior to the state-of-the-art methods.

## I. Introduction

**D**RIVEN by advances in aerospace and sensor technology, an increasing number of high-resolution remote sensing imagery (HRSI) can be captured globally, which have rich spatial details and potential semantic contents. Identifying the semantic category of each pixel in HRSI has become an important task of remote sensing intelligent interpretation in community. At present, the semantic segmentation of HRSI has been widely applied to urban planning [1-3], intelligent transportation [4, 5], disaster prediction [6-8], and agricultural production [9, 10].

In the past decade, deep learning methods, especially deep convolutional neural networks (DCNN), have shown excellent performance in semantic segmentation tasks of HRSI. Compared to traditional machine learning methods for semantic segmentation, such as support vector machines [11], random forests [12], conditional random fields [13] , etc., DCNN-based methods like FCN [14], U-Net [15], SegNet [16], Deeplab series [17-19], can capture multi-level and multi-scale semantic information and finer grained context information, demonstrating great potential in feature representation and pattern recognition. However, due to the limitation of fixed receptive fields in convolutional operation, DCNN-based methods lack the ability to model global contextual information and long-range dependencies. Even though self-attention based methods can alleviate the above problems [20, 21], they obtain global information based on the aggregation of local feature obtained by DCNN, making it difficult to obtain accurate global context information in complex HRSI scenes. Recently, inspired by the success of Transformer [22] in extracting global contextual relationships, Transformer-based methods have gradually been applied to semantic segmentation of HRSI [23-25], and have shown excellent performance in extracting long-range dependencies of geo-objects and global context information.

Most DCNN-based semantic segmentation models assumes that the data distribution of the training and testing sets is the same. However, different remote sensing image domains often face the problem of "data distribution gap" mainly due to the discrepancy in sensor type, imaging conditions, and shooting location. Facing this situation, traditional semantic segmentation models suffer from significant performance drops and fail to achieve out-of-distribution generalization [26, 27]. To address this problem, UDA methods aim to transfer the knowledge learned from the labeled source domain to the unlabeled target domain without extensive sample annotations

Manuscript received ; revised ; accepted ; date of publication ; date of current version. This work was supported by the National Key Research and Development Program of China under Grant 2020YFA0713503, the National Natural Science Foundation of China under Grant 42071427, 42371393, the Natural Science Foundation of Hunan Province, China under Grant 2023JJ30655, and the Major scientific and technological projects of Yunnan Province under Grant 202202AD080010. (Corresponding author: Jie Chen.)

Jingru Zhu, Ya Guo, Geng Sun and Jie Chen are with the School of Geosciences and Info-Physics, Central South University, Changsha 410083, China (e-mail: cj2011@ csu.edu.cn).
L. Hong is with the College of Tourism & Geography Science, Yunnan Normal University, Kunming 650500, Yunnan, China (hongliang20433@hotmail.com).

on the target domain.

Currently, some UDA works relying on feature alignment have achieved remarkable performance for semantic segmentation by minimizing adversarial learning losses [27-30] or contrastive learning loss [31] between source and target domains. Specifically, adversarial learning-based methods typically involve two-players: feature extractor and discriminator. The former uses a CNN-based network to extract features from images of source and target domains, while the latter determines whether the input features are from source or target domain. The model is well-trained when two-players reach a Nash equilibrium in a min-max game through adversarial learning, where the feature extractor can produce aligned and indistinguishable feature representations between two domains. In order to ensure the consistency of the local joint distribution when the alignment of global distribution, some adversarial learning-based works attempt to take the category-wise information into account during the adversarial learning. For example, a category-level adversarial network [29] was proposed to weight the adversarial loss for different features by emphasizing the role of category-level feature alignment in reducing domain shifts. Chen et al. [32] adaptively reduced the attention of classifier on category-level aligned features while increasing the attention on category-level unaligned features by considering category-certainty maps. FADA [33] proposed a fine-grained adversarial learning strategy to incorporate class information into the discriminator for aligning features at a fine-grained level. Unlike the adversarial learning, a contrastive adaptation framework [31] was proposed to simultaneously consider the alignment of intra-class features and the utilization of inter-class relationships.

These methods tend to align pixels or features based on learned statistical information. However, the statistical dependence based methods are prone to exhibit uncertainty and vulnerability in the prediction results due to the interference of chaotic phenomena [34]. Therefore, for better realizing the high-performance semantic segmentation between different domains of HRSI, merely considering the statistical dependency between variables is insufficient, but an underlying causal model [35-39]. For example, the tree in HRSI with RGB band presents green color, showing high statistical dependence between the "green color" and "tree" label in the semantic segmentation task of HRSI, which could easily mislead the model to make wrong prediction when the tree presents red color in HRSI with IRRG band in the target domain. After all, the essential characteristics of tree are the rounded crown, regular textural features in HRSI, instead of the color make a tree 'tree'.

In order to explore the invariant causal mechanism between different HRSI domains and semantic labels, in this paper, we address the cross-domain semantic segmentation task of HRSIs from a causal view and formalize it as structure causal model (SCM) [40] by examining on the causalities among the following six factors: unobserved causal factors $C$, unobserved non-causal factors $U_S$ and $U_T$, observed images $X_S$ and $X_T$, and ground truth labels $Y$. Specifically, we assume that the semantically relevant information in images $X_S$ and $X_T$ serves as causal factors, which are usually domain-invariant information, e.g., information about the content structure in images that corresponds to the semantic labels. And assume category-independent information as non-causal factors, which are usually domain-specific information, e.g., the style information in images. Fig. 1(a) illustrates the SCM of UDA semantic segmentation of HRSI, where $U_S \rightarrow X_S \leftarrow C$ ($U_T \rightarrow X_T \leftarrow C$) represents that each observed $X_S$ ($X_T$) is composed of domain-invariant $C$ and domain-specific $U_S$ ($U_T$). $U_S \leftarrow\!\cdots\!\rightarrow C$ ($U_T \leftarrow\!\cdots\!\rightarrow C$) denotes a pseudo-correlation between the $U_S$ ($U_T$) and $C$, which is usually caused by unobserved confounders. $X_S \rightarrow F_S \rightarrow Y$ and $X_T \rightarrow F_T \rightarrow Y$ represents that existing UDA methods usually learn the images embedding $F_S$ ($F_T$) based on the observed $X_S$ ($X_T$), and make the prediction $Y$ based on the $F_S$ ($F_T$). $C \rightarrow Y$ represents that the $C$ is the only parent variable that determines the generation of semantic label $Y$. However, existing UDA methods typically ignore the causal mechanism between source and target domains.

According to $d$-connection theory [41]: two variables are interdependent if they are connected by at least one unblocked path. We can find two paths that will lead to a pseudo-correlation between $U_S$ ($U_T$) and $Y$ in Fig. 1(a): 1) $U_S \rightarrow X_S \rightarrow F_S \rightarrow Y$ or $U_T \rightarrow X_T \rightarrow F_T \rightarrow Y$ and 2) $U_S \leftarrow\!\cdots\!\rightarrow C \rightarrow Y$ or $U_T \leftarrow\!\cdots\!\rightarrow C \rightarrow Y$. In order to make prediction $Y$ solely dependent on $C$, we need to intercept the above two unblocked paths. To this end, in order to intercept path 1), we hope to separate the potential domain-invariant $C$ and the domain-specific $U_S$ ($U_T$) from $X_S$ ($X_T$), and predict $Y$ solely based on $C$, namely $U_S \leftarrow X_S \rightarrow C \rightarrow Y$ or $U_T \leftarrow X_T \rightarrow C \rightarrow Y$. In order to intercept path 2), one possible solution is to make $C$ and $U_S$ ($U_T$) irrelevant as we cannot change the connection between $C$ and $Y$, namely $U_S \leftarrow\!\cdots\!\times\!\cdots\!\rightarrow C$ or $U_T \leftarrow\!\cdots\!\times\!\cdots\!\rightarrow C$. Fig. 1(b) illustrates the SCM of our proposed method. Unfortunately, since causal/non-causal factors are generally unobservable and unformulated, we cannot disentangle causal and non-causal factors directly from the input images [42, 43]. Therefore, in this paper, we introduce the disentangled representations learning to characterize the causal features, and make them satisfy the following three principles for replacing causal factors: 1) the causal factors should be separated from the non-causal factors, 2) the causal factors of source and target domains should be domain invariant, and 3) intervention on non-causal factors will not affect the causal factors to make decisions on labels.

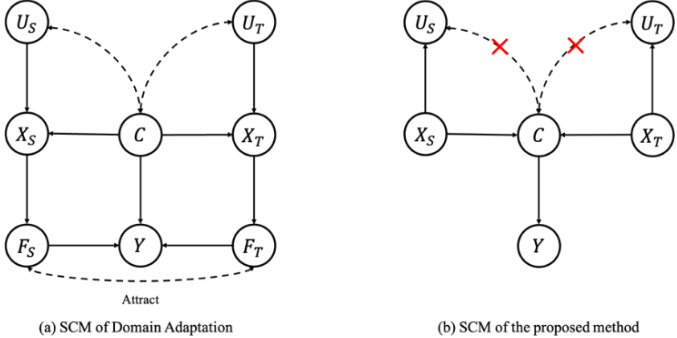

Fig. 1. SCM of UDA semantic segmentation of HRSI. The solid arrow indicates that the parent node causes the child one; while the dash arrow means there exists statistical dependence.

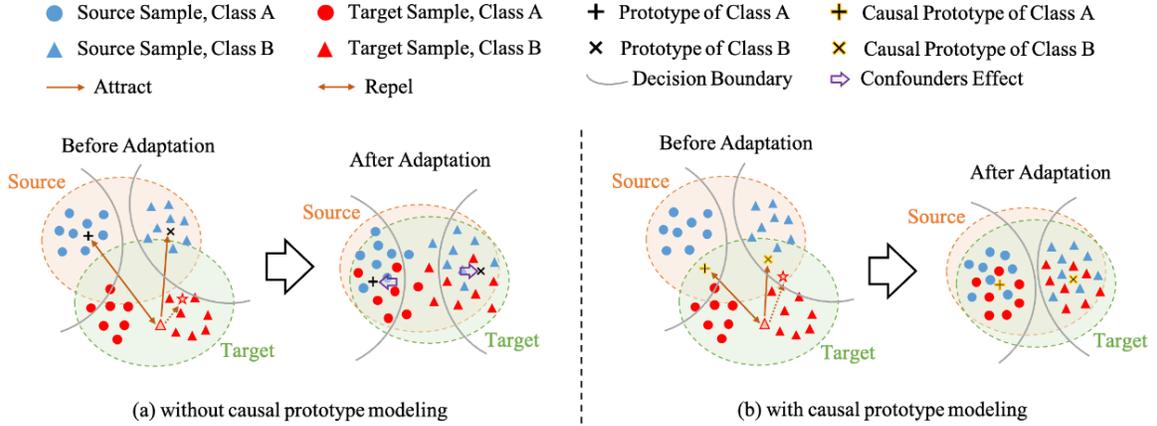

Fig. 2. Illustration of causal prototype modeling.

Based on the above analysis, we proposed a causal prototype-inspired contrast adaptation (CPCA) method for the UDA semantic segmentation task of HRSIs. Firstly, we design a causal feature disentanglement (CFD) module to disentangle the causal features and bias features from the original HRSIs for satisfying the first principle of causal factors. Notable, the current causal features are respectively independent of the source and target domains rather than domain-invariant. To further learn the domain-invariant causal features, we develop a causal prototypical contrast (CPC) module to bridge the discrepancy between source and target domain causal features, which can satisfy the second principle of causal factors. Finally, causal features should be predictively adequate for semantic labels, and its decision-making will not be affected when intervening on the bias features. Therefore, we introduce a causal intervention (CI) module to intervene on bias features to generate counterfactual unbiased samples for further decorrelating the causal and bias features, which can satisfy the third principle of causal factors. As shown in Fig. 2, compared with the traditional UDA model, our proposed model with causal prototype modeling can learn to characterize causal features during domain adaptation, which leads to a more compact domain-invariant feature space where features of the target domain appear correctly within the decision boundary.

The contributions of our work are as follows:

1) We introduce a causal view to formalize the cross-domain semantic segmentation problem for HRSI, and propose a CPCA algorithm to separate causal features from the original input HRSIs by disentangled representation learning, and force the causal features to replace causal factors by satisfying three principles that causal factors should possess in UDA task.

2) We propose a causal prototypical contrastive learning to bridge the discrepancy between the source and target domain causal features, motivating the model to obtain domain-invariant causal features.

3) A causal intervention module is proposed to exert intervention on bias features to generate counterfactual unbiased samples, which further de-correlates causal features and bias features so that causal features are sufficiently predictive for semantic labels.

The remainder of this study is organized as follows. Section II summarizes the related work to our study. Section III introduces the proposed method in detail. Section IV discusses the experiment settings, comparative experimental result analysis, ablation experiments, and visualization of causal features and bias features. Section V summarizes our work.

## II. RELATED WORK

### A. UDA for Semantic Segmentation

Existing methods for UDA semantic segmentation task can be divided into three groups, including image-to-image translation, adversarial learning, and self-training. The image-to-image translation methods aim to eliminate the style differences between source and target domains by using a generative adversarial network (GAN) framework to perform style transfer between source and target domains [44, 45]. Adversarial learning methods focus on adopting an adversarial loss on features space [27] or output space [30] to generate an invariant space shared by source and target domains. Some studies have reinforced the learning of invariant spaces by introducing category information [29, 32] or multiscale information [28] in the adversarial learning. Furthermore, [31] used contrastive learning instead of adversarial learning to align the intra-class distribution while distancing the inter-class distribution between the source and target domains. On the other hand, inspired by semi-supervised learning (SSL), self-training adaptative methods focus on assigning high-confidence pseudo-labels to the target domain and involving them in the training of semantic segmentation models. An iterative self-training methods [46] was proposed to iteratively generate pseudo-labels on target domains and re-training the semantic segmentation model with these labels. ProDA [47] introduced a prototypical pseudo label denoising method to correct the soft pseudo labels of target domain, enabling the network to learn from high-confidence denoised pseudo-labels of target domain during training.

In the field of remote sensing, the performance of semantic segmentation models is usually affected by the differences in sensor types, imaging conditions, imaging regions, etc. To solve this problem, domain adaptation methods have been gradually applied to the semantic segmentation task of HRSIs. To consider the consistency of local distribution along with the alignment of global distribution, an category-certainty attention is proposed to focus on category-level unaligned regions in remote sensing images [32]. [48] introduced an intermediate domain to connect source and target domains, and used it to achieve knowledge transfer between the source domain and the target domain. MemoryAdaptNet [49] introduced an invariant feature memory module to learn the domain-invariant features, and aggregate it into the current image to further enhance the representations of the current features by employing a category attention-driven aggregation module. A multi-prototypes learning methods [50] was proposed to model complex intra-class and inter-class relationships for effectively alleviating the effect caused by intra-domain and inter-domain differences in UDA semantic segmentation of HRSIs.

However, these current UDA semantic segmentation models are committed to learning the statistical information related to semantic labels from source and target domain images, and aligning or transforming distributions based on it to make predictions for target domain images. Obviously, these models are susceptible to the influence of uncertain factors, and show vulnerability and non-interpretability in predicting labels when facing complex and variable remote sensing image scenes. In contrast, we adopt a UDA semantic segmentation method of HRSI from a causal view with structure casual model, which focuses on exploring the inherent causal mechanism of cross-domain remote sensing images by learning domain-invariant causal features to provide better generalization ability and interpretability.

*B. Causal Learning*

Statistical learning models can learn the correlation of data. It can perform better in independent and identically distributed (i.i.d.) setting when enough i.i.d. data are observed. However, these methods typically perform poorly when the data does not meet the i.i.d condition or when the modeling scenery is complex. In contrast, causal learning [41] strives to discover more essential causal structures from the data beyond correlations. According to Judea Pearl's causal ladder [51], causal reasoning can be divided into three levels: association, intervention and counterfactual, which can describe the variant and invariant parts of the data distribution by modeling the human processes of observation, action, decision-making, and reflection. Therefore, over the past few years, causal learning has received increasing attention in the areas of data generation, domain generalization, and model interpretability [52, 53]. For example, a graph surgery estimator relying on causal directed acyclic graph (DAG) was proposed for recognizing and representing the invariance required for changes due to dataset shifts [54]. [55] viewed the data generating process from a causal perspective and categorized the latent data generating factors into invariant and style features. For the domain generalization task, MatchDG [55] proposed a structural causal model to formalize the domain generalization problem and illustrated the importance of intra-class variation modeling for generalization. CIRL [43] exploited dimension-wise representations to mimic causal factors based on theoretical properties and then reconstructed invariant causal mechanisms to address domain generalization. For explaining black-box visual classifiers, [56] formulated a causal extension to the paradigm of instance-wise features selection and obtain a subset of input features that has the greatest causal effect on the model's output. [57] conducted a comprehensive review of existing causal reasoning methods for visual representation learning and pointed out the importance of causal reasoning in visual representation learning.

These studies are usually based on causality to obtain invariant causal mechanisms or recover causal features, thereby improving the generalization problem of out-of-distribution (OOD) or the interpretability problem of visual representation learning. However, few existing methods are designed for UDA semantic segmentation of HRSIs. Since the observable variables and latent causal variables in structural causal model are different between UDA task and DG task, it is necessary to design causal structure model for the task of UDA semantic segmentation of HRSIs, and to mine the causal invariant mechanism of HRSIs in cross-domain scenarios.

*C. Disentangled Representation Learning*

Disentangled representation learning (DRL) aims to identify and reveal hidden representations in observational data with human-like generalization capabilities, attempting to break the black-box properties of existing end-to-end deep learning models. Existing studies [58, 59] have demonstrated the potential of DRL in modeling human learning and understanding of the world, mainly because DRL encourages the representations to carry interpretable semantic information with independent factors, and shows significant potential for representing invariance [60, 61], integrity [62, 63], and generalization [64, 65]. In view of this, CRL is now widely applied in computer vision [66-70], natural language processing [71-75], and graphical learning [76-78]. The most typical approaches of DRL are based on generative models such as variational auto-encoder (VAE) [79] or generative adversarial network (GAN)[80]. For example, IDVAE [81] designed a new identifiable double VAE framework for learning optimal representation of latent space to learn disentangled representations. A disentangled-VAE [65] was proposed to disentangle category-distilling and category-dispersing factors from visual and semantic features, respectively, for avoiding non-recognition related information in zero-shot learning. Aiming to address age-invariant face verification task, DR-RGAN achieves disentangled representation by representing the facial features without age interference. These approaches show greater potential for learning interpretable representations of images. In addition, causal inference based DRL are also widely used. For example, a VAE-based causal disentangled representation learning framework named CausalVAE [62] was proposed to recover the latent factors with semantics and structure via a causal DAG. [82] presented a causal perspective for learning trustworthy and transferable disentangled representations, and captured the generation process of latent variable representation learning models. These approaches aim to encourage network to learn potential causal disentangled representations from a causal perspective. Inspired by this, we propose a disentangled representation learning method for better solving the domain shift problem of UDA semantic segmentation of HRSI by disentangling domain-invariant causal features from source- and target-domain images.

## III. METHODOLOGY

In this section, we focus on the causal prototype-inspired UDA semantic segmentation methods for HRSIs. Given the source domain images $X_S$ with $K$-class pixel-level labels $Y_S$ and the target domain images $X_T$ without pixel-level labels. Our goal is to construct domain-invariant causal mechanisms from labeled source domain images and unlabeled target domain images based on above three principles of causal factors, so that the model can be well generalized in the target domain. As show in Fig. 3, our proposed CPCA consists of three modules: causal feature disentanglement (CFD) module, causal prototypical contrast (CPC) module, and causal intervention (CI) module. Firstly, $X_S$ and $Y_S$ are input into CFD module to obtain causal features $C_S$, bias features $B_S$ of the source domain and causal features $C_T$, bias feature $B_T$ of the target domain, respectively. Secondly, for obtaining domain-invariant causal features between source and target domains, a CPC module is proposed to bridge the discrepancy between $C_S$ and $C_T$ by contrastive learning between $C_S$ ($C_T$) and causal prototypes. Finally, after the domain-invariant causal features are well-learned, we utilize the CI module to swap interventions on $B_S$ and $B_T$ in each min-batch to get counterfactual unbiased samples, which can enforce the classifier to learn the unbiased information and promote the predictive adequacy of causal features for semantic labels.

### A. Causal Feature Disentanglement Module

In order to eliminate the spurious correlation between bias factors $B_S$ ($B_T$) and label $Y$, and intercept the paths 1) and 2) in SCM of Fig. 1(a), we need to disentangle potential causal and non-causal factors from the observed images $X_S$ and $X_T$. However, since both causal and non-causal factors are hidden variables rather than directly observed variables, it is difficult to directly disentangle the images $X_S$ and $X_T$ into $X_S = f(C, U_S)$ and $X_T = f(C, U_T)$, where $f(\cdot)$ is the disentanglement function. Inspired by recent works on disentangled representation learning [82-84], in this paper, we disentangle the causal and bias features from the observed images from the perspective of disentangled representation learning, and attempts to substitute causal factors with causal features that force them to satisfy the three principles of causal factors.

In this section, we introduce the CFD module to disentangle causal and bias features for carrying out the separation of causal features from bias features. Specifically, as shown in Fig.3, the CFD module consists of a causal encoder $E_c$, and bias encoder $E_b$, which are used to extract causal and bias features from input images, respectively. Given a mini-batch $X_S$ and $X_T$, we input $X_S$ and $X_T$ into $E_c$ to produce source causal features $C_S$ and target causal features $C_T$. Meanwhile, the $X_S$ and $X_T$ are inputted into $E_b$ simultaneously, and mapped into bias features $B_S$ and $B_T$. $E_c$ and $E_b$ share the same network structure, and can be any DCNN-based network. In this paper, we use ResNet-50 [85] pretrained on source dataset as the backbone of $E_c$ and $E_b$. Finally, through the CFD module, we disentangle the $X_S$ into $C_S$ and $B_S$, and disentangle the $X_T$ into $C_T$ and $B_T$. Intuitively, $E_c$ and $E_b$ can focus on the different part of structure information of original image.

### B. Causal Prototypical Contrast Module

The $C_S$ and $C_T$ obtained by CFD module are independent of the source and target domains, respectively. To further achieve better generalization in the target domain, the causal features should be domain-invariant. Therefore, in order to learn domain-invariant causal features so that the causal features satisfy the second principle of causal factors, in this section, we use the CPC module to construct category-level domain-invariant causal prototypes $p$, and bridge the discrepancy between the $C_S$ ($C_T$) and the $p$ through contrastive learning. Specifically, CPC module consists of three steps: causal prototypes initialization, causal prototypes contrastive learning, and causal prototypes updating.

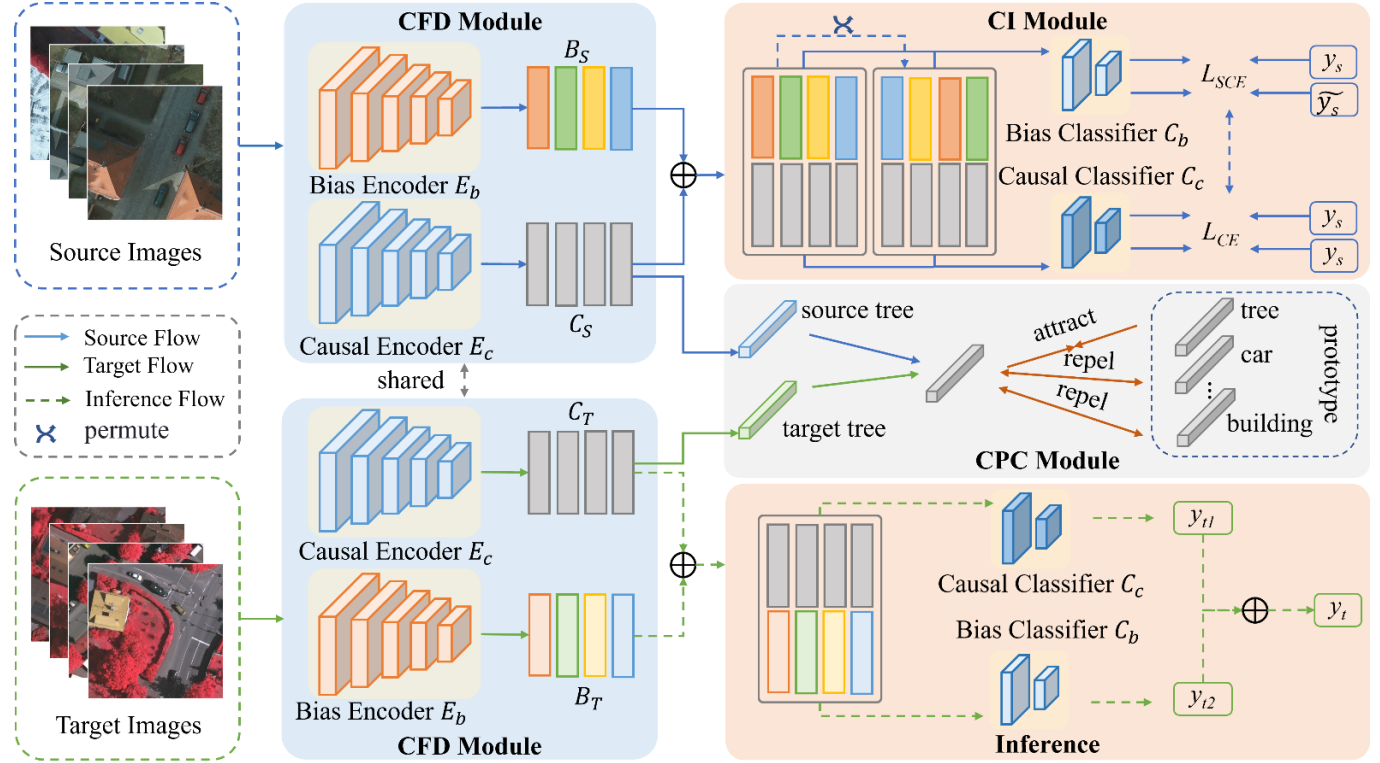

Fig. 3. Structure of CPCA.

1） *Causal prototypes initialization*

Before domain adaptation, we first train the semantic segmentation model on the labeled source domain dataset in a fully supervised manner and use the cross-entropy loss function to constraint the training of segmentation model, which is shown as follows:

$$L_{seg}(X_S, Y_S) = -E_{(x_s,y_s)\sim(X_S,Y_S)} \sum_{i=1}^{K} y_s^{(i)} log \hat{y}_s^{(i)}, \quad (1)$$

where $L_{seg}(X_S, Y_S)$ is the segmentation loss, $x_s \in X_S$ represents the images from source domain, and $y_s \in Y_S$ represents the ground truth corresponding to $x_s$, $\hat{y}_s$ represents the predicted label corresponding to $x_s$, and $K$ represents the number of classes.

After obtaining the model trained on the labeled source domain, we initialize the $p$ with deep features of the source domain:

$$p = \left\{ p_1, p_2, \dots, p_K \middle| p_k = \frac{\sum_{n=1}^{N_S} C_n^S \mathbb{I}[Y_n^S = k]}{\sum_{n=1}^{N_S} \mathbb{I}[Y_n^S = k]} \right\}. \quad (2)$$

where $N_S$ is the number of source domain images, $C_n^S \in R^d$ is the source causal feature vector with dimension d, $k$ is the index of categories number $K$, $\mathbb{I}[Y_n^S = k]$ is an indicator function, which equals to 1 if $Y_n^S = k$ and 0 otherwise.

2） *Causal prototypes contrastive learning*

In order to bridge the discrepancy between $C_S$ and $C_T$ and obtain domain-invariant causal features, we use the contrastive learning paradigm that considers both inter-class and intra-class relationships instead of adversarial learning paradigm [27, 28, 30] for feature alignment in this paper. Firstly, we compute the similarity $S_S^k$ ($S_T^k$) between $C_S$ ($C_T$) and causal prototype $p_k$ of each category $k$, respectively:

$$S_S^k = \frac{\exp(p_k \cdot C_S / \tau)}{\sum_{k=1}^{K} \exp(p_k \cdot C_S / \tau)}, \quad (3)$$

$$S_T^k = \frac{\exp(p_k \cdot C_T / \tau)}{\sum_{k=1}^{K} \exp(p_k \cdot C_T / \tau)}, \quad (4)$$

where $\tau$ represents the temperature that controlling the sharpness of generated probability distribution. Then, we calculate the cross-entropy loss between the $S_S^k$ ($S_T^k$) and semantic labels:

$$L_S = -\sum_{k=1}^{K} y_S^k \log S_S^k, \quad (5)$$
$$L_T = -\sum_{k=1}^{K} \tilde{y}_T^k \log S_T^k, \quad (6)$$

where $y_S^k$ is the ground-truth label of source domain, $\tilde{y}_T^k$ is the predicted label of target domain. The final causal prototypes contrastive loss is:

$$L_{CC} = L_S + L_T. \quad (7)$$

CPC module can narrow the discrepancy of causal features in the same category, while widening the discrepancy of causal features in different categories between source and target domains, thus achieving the learning of domain-invariant causal representation.

*3) Causal prototypes updating*

To enhance the domain-invariant representation of causal prototypes, we integrate the causal information of source and target domains into $p$ with a moving average scheme. Specifically, we first update the causal features $C_k$ for each category $k$ based on $S = \{S_S^k, S_T^k\}$ as:

$$C'_k = \sum_{i=1}^{L} \frac{1 - S^i}{\sum_{j=1}^{L}(1 - S^j)} \cdot C_k^i, \quad (9)$$

where $L$ represents the number of pixels labeled as $k$ in $C$, $C_k^i = \{C \mathbb{I}[Y = k]\} \in C_k$.

Finally, we calculate the updated prototype of each category $k$ by leveraging moving average:

$$p'_k = (1 - m) \cdot p_k + m \cdot C'_k, \quad (10)$$

where $m$ is the momentum, and we employ polynomial annealing policy to update it:

$$m_t = (1 - \frac{t}{T})^\alpha \cdot \left(m_0 - \frac{m_0}{100}\right) + \frac{m_0}{100}, t \in [0, T]. \quad (11)$$

where $m_t$ represents the value of $m$ at the $t_{th}$ iteration, $T$ represent the total number of training iterations, and $\alpha$ and $m_0$ are default parameters and set to 0.9.

*C. Causal Intervention Module*

According to above description, the causal and bias features from source and target domain images can be disentangled using the CFD module, and the causal features of source and target domains can conform to the domain-invariant principle using the CPC module. However, the causal features $C_S$ ($C_T$) and the bias feature $B_S$ ($B_T$) are disentangled from the biased source domain (target domain) images, and there is a statistical correlation between $C_S$ ($C_T$) and $B_S$ ($B_T$). Given the Judea Pearl's Ladder of causality theory[41]: the causal factors should satisfy the requirements of intervention and counterfactuals, we argue that the causal features $C_S$ ($C_T$) should remain **invariant** to interventions on $B_S$ ($B_T$), i.e., $P(C_S | do(B_S))$ or $P(C_T | do(B_T))$, when causal features and bias features are not confounded.

Therefore, to further decorrelate $C_S$ ($C_T$) with $B_S$ ($B_T$), in this paper, we intervene on the $B_S$ by swapping in each min-batch to generate counterfactual instances. More specifically, for the bias features $B_S$ of each min-batch, we perform random permute at the min-batch to generate the permuted bias features $\widehat{B_S}$. Then, we concatenate it with the causal features $C_S$ in the same min-batch to obtain the counterfactual features $F_S^{unbiased} = [C_S; \widehat{B_S}]$.

Since $C_S$ and $\widehat{B_S}$ in $F_S^{unbiased}$ may be random combination from different images, they have less correlation than $C_S$ and $B_S$ in $F_S^{biased} = [C_S; B_S]$ where both from same images. Based on the principle that the causal factors should remain predictive adequacy for interventions on non-causal factors: classification of the $F_S^{unbiased}$ should make the same prediction as the pre-intervention features $F_S^{biased}$. Therefore, for the $F_S^{biased}$ and $F_S^{unbiased}$, we use the following loss function to train the causal classifier $C_c$:

$$L_C = L_{C_1} + L_{C_2}, \quad (12)$$
$$L_{C_1} = CE\left(C_c\left(F_S^{biased}\right), y_s\right), \quad (13)$$
$$L_{C_2} = CE\left(C_c\left(F_S^{unbiased}\right), y_s\right), \quad (14)$$

where $CE(.)$ represents the cross-entropy loss, $y_s$ represents the source label.

Meanwhile, to ensure that bias encoder $E_B$ focus on the bias information, and decorrelation $C_S$ and $B_S$, we further swap label $y_s$ as $\widehat{y_s}$ along with $\widehat{B_S}$ to train the bias classifier $C_b$:

$$L_B = L_{B_1} + L_{B_2}, \quad (15)$$
$$L_{B_1} = SCE(C_b(F_S^{biased}), y_s), \quad (16)$$
$$L_{B_2} = SCE(C_b(F_S^{unbiased}), \hat{y}_s), \quad (17)$$

where $SCE(.)$ represents the symmetric cross entropy [86].

Together with the causal prototypes contrast loss, total loss function is defined as:

$$L = L_{CC} + L_C + L_B. \quad (18)$$

It is worth noting that in order to ensure the quality of the counterfactual features $\widehat{B_S}$ and $\widehat{B_T}$, in the early stage of training, we only use $L_{CC}$, $L_{C_1}$, and $L_{B_1}$ to train the model. After $\tau$ iterations, we use $L$ to train the model.

*D. Self-training*

After the causal-prototype contrast adaptation stage, we can further improve the performance of UDA semantic segmentation of HRSI by self-training strategy following previous works[87]. Given the target images $X_T = \{x_n^t\}_{n=1}^{N^t}$, where $N^t$ is the number of target images. We input each image $x_t^n$ to the causal-prototype contrast adaptation model, and to obtain the predicted label set $\tilde{Y}_T = \{\tilde{Y}_n^t \in R^{H \times W}\}_{n=1}^{N^t}$ and a ranked confidence set $\delta_k = [\delta_k^1, \delta_k^2, \cdots, \delta_k^l]$ of category $k$, where $l$ represents the length of confidence set belongs to category $k$, it can be calculated as follows:

$$l = \sum_{n=1}^{N_t} \sum_{i=1}^{H} \sum_{j=1}^{W} \mathbb{I}[\tilde{Y}_n^t = k], \quad (19)$$

where $\tilde{Y}_n^t$ is the predicted label for the image $x_n^t$, $\mathbb{I}[\tilde{Y}_n^t = k]$ is an indicator function, which equals to 1 if $\tilde{Y}_n^t = k$ and 0 otherwise. Then, we use threshold $\sigma_k$ to determine whether a pixel is associated with category $k$:

$$\tilde{P}_n^t = \{\tilde{Y}_n^t \in R^{H \times W} | \delta_k \geq \sigma_k\}. \quad (20)$$

where $\tilde{P}_n^t$ represents the pseudo label for the image $x_n^t$, $\sigma_k = l * \eta$, $\eta$ is the percentage.

## IV. EXPERIMENTS AND RESULT ANALYSIS

*A. Experimental Settings*

*1) Datasets*

The data domain gap between HRSIs is mainly reflected in the differences in color texture, scale size, and shape characterization of geo-object classes caused by the differences in sensor type, resolution size, imaging location, or imaging weather. In order to verify that our method can better handle these common distribution differences of HRSIs in semantic segmentation tasks, we conduct experiments on two HRSI datasets provided by International Society for Photogrammetry and Remote Sensing (ISPRS) WG II/4: Potsdam dataset [88] and Vaihingen dataset [89]. Specifically, both datasets provide a certain number of high-resolution true orthophoto tiles (TOPs) with their corresponding pixel-by-pixel semantic labels, which consist of six general land cover categories: Imp. Surf. (impervious surfaces), Building (buildings), Low Veg. (low vegetation), Tree (trees), Car (cars), and Clutter (clutter/background). In addition, there are data distribution differences in terms of sensor type, resolution size, and imaging location between these two datasets. For example, due to the sensor type is different, the Potsdam dataset provides four bands: near-infrared (IR), red (R), green (G), and blue (B), while Vaihingen dataset only provides three bands: the near-infrared (IR), red (R), and green (G), so we can select three bands (such as R, G and B bands) from the Potsdam dataset that are different from the Vaihingen dataset and combine them to simulate the domain differences under different sensor types. In terms of resolution size, the Potsdam dataset contains 38 TOPs with a size of 6,000 × 6,000 and a resolution of 5 cm, and the Vaihingen dataset contains 33 TOPs with a size of approximately 2,494 × 2,064 and a resolution of 9 cm. In terms of imaging locations, the Potsdam dataset describes an urban scene with a crowded residential pattern, while the Vaihingen dataset describes a small rural scene with a sparse layout. Therefore, setting up a cross-domain semantic segmentation task between these two datasets can fully test the ability of our method to handle the common domain differences in remote sensing image.

*2) Task Settings*

In this paper, we set up three UDA semantic segmentation tasks: POT(IR-R-G)→VAI (IR-R-G), VAI (IR-R-G)→POT (IR-R-G), and POT (R-G-B)→VAI (IR-R-G), where POT (IR-R-G)→VAI (IR-R-G) represents the Potsdam dataset with a band combination of IR-R-G as the source domain and the Vaihingen dataset with a band combination of IR-R-G as the target domain; VAI (IR-R-G) →POT (IR-R-G) represents the Vaihingen dataset with a band combination of IR-R-G as the source domain and the Potsdam dataset with a band combination of IR-R-G as the target domain; POT (R-G-B) → VAI (IR-R-G) represents the Potsdam dataset with a band combination of R-G-B as the source domain and the Vaihingen dataset with a band combination of IR-R-G as the target domain. The domain discrepancy information on these three UDA semantic segmentation tasks is shown in TABLE I.

We crop the Potsdam dataset and the Vaihingen dataset into image patches according to the size of 512*512 and the step size of 256 in the model training phase where the training set of the Potsdam dataset contains 11,616 patches, and the test set contains 1,694 patches; the training set of Vaihingen dataset contains 1,296 patches, and the test set contains 112 patches. The rest of the detailed settings are the same as in [49]. In particular, the source domain images with their semantic labels, and the target domain images are involved during training, while the semantic labels of the target domain are not accessible during training.

TABLE I
THE DOMAIN GAPS BETWEEN THREE UDA TASKS.

| Domain gaps | Datasets | | | Tasks | | |
| --- | --- | --- | --- | --- | --- | --- |
| | POT(IR-R-G) | POT (R-G-B) | VAI (IR-R-G) | POT(IR-R-G) → VAI (IR-R-G) | POT (R-G-B) → VAI (IR-R-G) | VAI (IR-R-G) → POT (IR-R-G) |
| Spatial resolution (m) | 0.05 | 0.05 | 0.09 | √ | √ | √ |
| Geographic location | Potsdam | Potsdam | Vaihingen | √ | √ | √ |
| Scene | city | city | small village | √ | √ | √ |
| Imaging sensors | IR-R-G | R-G-B | IR-R-G | | √ | |

*3) Implementation Details*

We implement our network using the PyTorch toolbox on a single RTX 3090 GPU with 24-GB memory. All experiments were implemented in the PyCharm Development Environment. The key parameters for network training are shown in Table II.

TABLE II
THE KEY TRAINING PARAMETERS INVOLVED IN THIS WORK

| Parameters | Source-only model | CPCA model | Self-training model |
|---|---|---|---|
| Optimizer | SGD | SGD | SGD |
| Initial learning rate | $5 \times 10^{-4}$ | $1 \times 10^{-3}$ | $5 \times 10^{-4}$ |
| Learning rate momentum | 0.9 | 0.9 | 0.9 |
| Learning rate weight decay | $5 \times 10^{-4}$ | $5 \times 10^{-4}$ | $5 \times 10^{-4}$ |
| Batch size | 8 | 2 | 8 |
| Training times of iteration | 20000 | 50000 | 20000 |

*4) Evaluation Metric*

We use three commonly evaluation metrics: $F_1$ score, OA (over accuracy), and mean IoU (Intersection and Union) to evaluate the performance of different UDA semantic segmentation methods.

Specifically, $F_1$ score is formulated as shown in Equation as follow:

$$F_1 = 2 \times \frac{Precision * Recall}{Precision + Recall}, \quad (21)$$

where the *Precision* and *Recall* can be formulated as follows:

$$Precision = \frac{TP}{TP+FP}, \quad (22)$$

$$Recall = \frac{TP}{TP+FN}, \quad (23)$$

Moreover, *OA* is as shown in Equations (24):

$$OA = \frac{TP}{TP+FP}, \quad (24)$$

For mIoU, we first compute the IoU for each category separately and then average them to obtain mIoU. The Equations (25) shows the equation to compute the IoU for each category.

$$IoU = \frac{TP}{TP+FP+FN}. \quad (25)$$

where *TP* denotes the number of true positive pixels, *FP* denotes the number of false positive pixels, and *FN* denotes the number of false negative pixels, respectively.

### B. Comparative Experimental Result Analysis

For verifying the validity of the proposed method, we perform comparison experiments on a Source-only model without UDA and eight state-of-the-art UDA methods without considering the causality (MCD_DA, AdaptSegNet, AdvEnt, FADA, SDCA, MUCSS, Zhang's and ProCA), in which MUCSS, Zhang's are the state-of-the-art UDA methods proposed in the remote sensing semantic segmentation, and we directly used the quantitative evaluation results reported in the original paper.

MCD_DA [90]: a UDA method that attempts to align source and target domain distributions by considering task-specific decision boundaries.

AdaptSegNet [28]: a UDA method employing multilevel adversarial learning in the output space for pixel-level semantic segmentation.

AdvEnt [30]: a UDA method for semantic segmentation via adversarial training approach based on the entropy of semantic predictions.

FADA[91]: a UDA method for class-level feature alignment by fine-grained adversarial learning strategy.

SDCA [92]: a UDA method that aligns the pixel-wise representation guided by semantic distribution.

MUCSS [93]: a UDA method under multiple weakly-supervised constraints to minimize the data shift of remote sensing images.

Zhang's [94]: a UDA method exploiting curriculum-style and local-to-global adaptation to address the cross-domain problem of remote sensing images.

ProCA [31]: a UDA method that aligns the class-centered distribution by prototypical contrastive learning.

All the comparative methods use the DeepLab-v3 with ResNet-101 pretrained on ImageNet as the semantic segmentation network. Limited by the Memory of GPU, our model uses the DeepLab-v3 with ResNet-50. All methods carry out experiments on these three tasks: POT(IR-R-G) →VAI (IR-R-G), VAI (IR-R-G) →POT (IR-R-G), and POT (R-G-B) → VAI (IR-R-G).

*1) Comparative Studies on POT (IR-R-G) →VAI (IR-R-G)*

The quantitative evaluation results of the compared domain adaptation methods on POT(IR-R-G) →VAI (IR-R-G) task are presented in Table III. From the results, due to the domain shifts in spatial resolution, geographic location and scene, the Source-only model has the worst performance with OA, MA, and mIoU values of 67.19%, 60.55%, and 43.58%, respectively. The eight state-of-the-art UDA comparison methods achieve varying degrees of performance improvement compared to the Source-only model due to accessing images in the target domain and using a domain-adaptive strategy. However, these methods tend to align pixels or features based on the statistical information learned from images and their corresponded labels, which leads to the uncertainty and non-interpretability of the prediction results when facing complex and variable remote sensing image scenes, limiting the performance of UDA semantic segmentation for HRSIs. Therefore, the best performing model ProCA among the comparison methods only achieves the performance with OA, MA, and mIoU values of 77.48%, 69.25%, and 54.37%, respectively. Considering the invariant causal mechanism between different remote sensing image domains and semantic labels in the process of domain adaptation, our method CPCA achieves the best performance with OA, MA, and mIoU values of 80.18%, 76.94%, and 60.75%, respectively. Compared with the source-only model, the CPCA shows the performance improvement of 12.99% on OA, 16.39% on MA, and 17.17% on mIoU. Compared with the comparison method ProCA with the best performance, the CPCA presents the improvement of 2.7%, 7.69%, and 6.38%

TABLE III
QUANTITATIVE EVALUATION RESULTS (%) OF DIFFERENT UDA MODELS ON THE POT(IR-R-G) →VAI (IR-R-G) TASK

| Methods | | $F_1$ score | | | | | | OA | MA | mIoU |
|---|---|---|---|---|---|---|---|---|---|---|
| | | Imp. surf. | Building | Low veg. | Tree | Car | Clutter | | | |
| Source-only | | 72.85 | 74.40 | 56.42 | 67.96 | 46.76 | 28.49 | 67.19 | 60.55 | 43.58 |
| State-of-the-art methods | MCD_DA (2018) | 77.00 | 80.30 | 54.94 | 70.48 | 54.28 | nan | 70.20 | 57.66 | 46.63 |
| | AdaptSegNet (2018) | 77.69 | 85.99 | 60.97 | 74.06 | 52.01 | 26.43 | 73.70 | 64.00 | 48.65 |
| | AdvEnt (2019) | 77.55 | 84.42 | 52.93 | 74.23 | 53.98 | 30.51 | 74.43 | 65.29 | 51.37 |
| | FADA (2020) | 78.77 | 87.84 | 60.84 | 78.31 | 57.82 | 37.76 | 76.60 | 67.72 | 52.55 |
| | SDCA (2021) | 81.80 | 88.55 | 56.47 | 77.29 | 64.19 | 31.20 | 76.71 | 67.77 | 52.79 |
| | MUCSS (2021) | 66.13 | 76.77 | 55.97 | 73.14 | 51.09 | 45.65 | 61.43 | - | 45.38 |
| | Zhang's (2022) | 80.13 | 86.65 | **64.16** | 71.90 | 61.94 | 31.34 | 66.02 | - | 52.03 |
| | ProCA (2022) | 82.42 | 90.31 | 63.68 | 78.49 | 56.54 | 37.45 | 77.48 | 69.25 | 54.37 |
| (Ours CPCA) | | **84.92** | **91.76** | 61.99 | **79.75** | **68.10** | **60.18** | **80.18** | **76.94** | **60.75** |

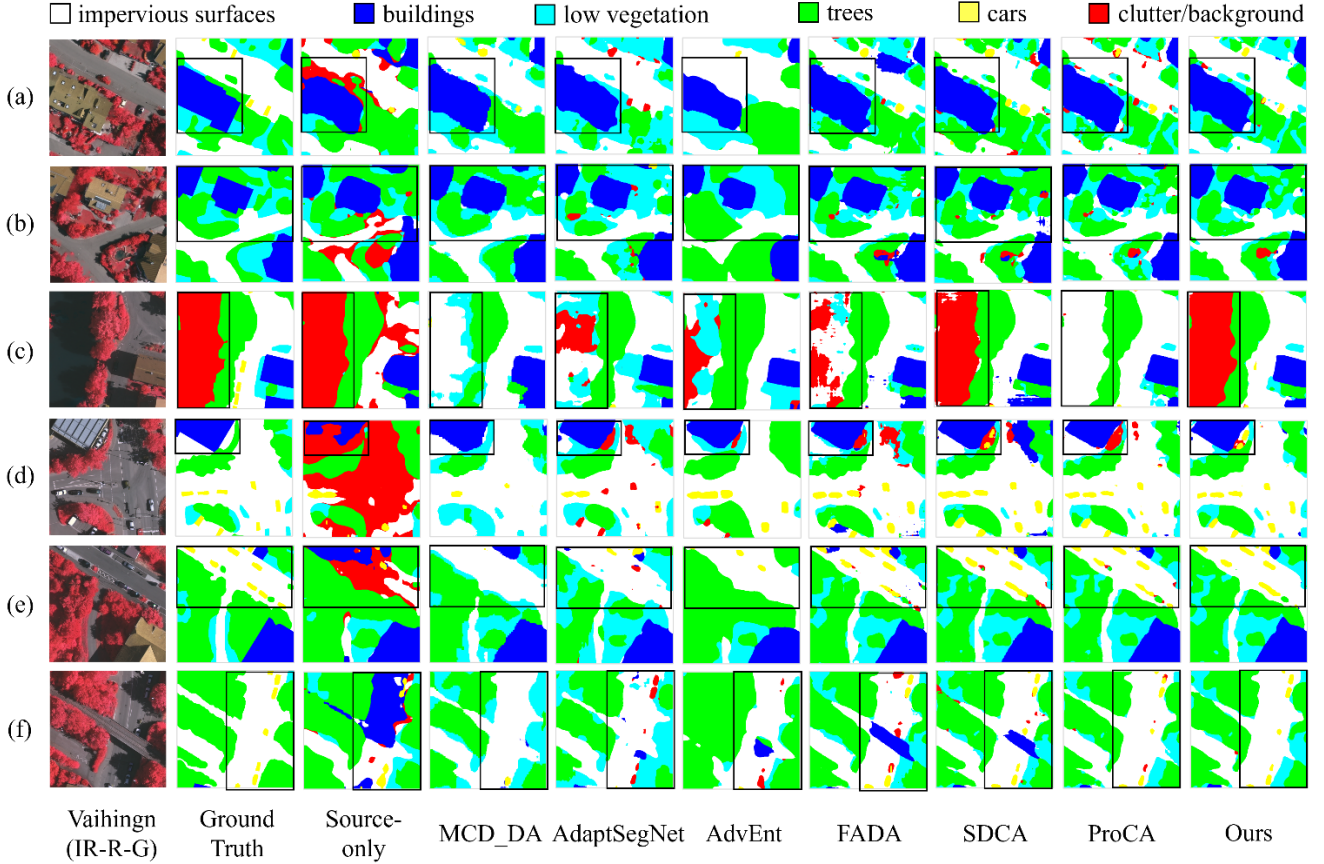

Fig. 4. Qualitative results on the POT(IR-R-G) →VAI (IR-R-G) task.

on OA, MA, and mIoU, respectively.

In Table III, the $F_1$ scores of CPCA in Imp. Surf., Building, Tree, Car and Clutter realize the best accuracies of 84.92%, 91.76%, 79.75%, 68.10% and 60.18%, respectively, and achieve the improvement of 12.07%, 17.36%, 11.79%, 21.34% and 31.69% compared to the Source-only model. Notable, the CPCA achieved the highest improvement of 60.18% compared to the Source-only model and 22.73% compared to the best performing model ProCA on the Clutter which contains multiple geo-objects such as water bodies, containers, tennis courts and swimming pools. This result demonstrates that the CPCA has higher performance and better generalization for such categories with complex features, multiple semantics and diverse scenes, which benefits from the fact that the CPCA mitigates the interference of complex appearance features and irrelevant external factors on the performance of semantic segmentation by capturing the invariant causal mechanism between images and semantic categories in different remote sensing image domains.

We selected six typical remote sensing scenarios in the test set of Vaihingen dataset for comparing the cross-domain semantic segmentation performance on the POT(IR-R-G) → VAI (IR-R-G) task. Fig. 4 is the qualitative results, where the semantic segmentation performance of Source-only model is poor due to the domain discrepancy between source and target domains, e.g., there is serious confusion between Clutter and other category, between Low. veg. and Tree, and between Imp. surf. and Building. After domain adaptation, the confusion problem has been improved, but there is still confusion between Clutter and Imp. surf., Car. By mining the causal relationship between two domain images and semantic labels, compared with other methods, our method achieves better semantic segmentation performance and improves the confusion problem between categories, such as between Clutter and Building, Low.veg., Tree in (a)(b); between Clutter and Imp. surf. in (c), between Clutter and Car in (d)(e)(f). In summary, compared to other UDA semantic segmentation methods without considering the causality, our method achieves the best semantic segmentation performance in the POT(IR-R-G) → VAI (IR-R-G) task.

*2) Comparative Studies on POT (R-G-B) →VAI (IR-R-G)*

The quantitative evaluation results of the compared domain adaptation methods on POT (R-G-B) →VAI (IR-R-G) task are presented in Table IV. Except for domain shift in spatial resolution, geographic location and scene layout, the discrepancy on POT (R-G-B) → VAI (IR-R-G) task also include imaging sensors. Due to these domain shift, the Source-only model also has the worst performance with OA, MA, and mIoU values of 57.38%, 48.07%, and 33.58%, respectively. By considering the narrowness of domain shift between the source and target domains, eight comparative state-of-the-art UDA methods all achieve better segmentation performance than the Source-only model, but there is still significant room for improvement. The proposed CPCA achieves the best performance with MA, and mIoU values of 71.75%, and 47.67%, respectively. Compared with the source-only model, our CPCA shows the performance improvement of 12.38% on OA, 23.68% on MA, and 14.09% on mIoU. Compared with the comparison method ProCA with the best performance, CPCA presents the improvement of 7.68%, and 2.04% on MA and mIoU, respectively. This demonstrates that feature alignment methods based on causal invariance are more effective in narrowing the distribution discrepancy between different domains.

TABLE IV
QUANTITATIVE EVALUATION RESULTS (%) OF DIFFERENT UDA MODELS ON THE POT (R-G-B) →VAI (IR-R-G) TASK

| | Methods | $F_1$ score | | | | | | OA | MA | mIoU |
|---|---|---|---|---|---|---|---|---|---|---|
| | | Imp. surf. | Building | Low veg. | Tree | Car | Clutter | | | |
| | Source-only | 64.91 | 63.58 | 43.54 | 63.97 | 40.44 | 7.94 | 57.38 | 48.07 | 33.58 |
| | MCD_DA (2018) | 69.75 | 75.46 | 51.62 | 50.85 | 35.05 | 0.00 | 61.75 | 48.30 | 36.52 |
| | AdaptSegNet (2018) | 68.31 | 78.23 | 46.27 | 63.73 | 40.44 | 8.09 | 63.13 | 51.08 | 37.55 |
| | AdvEnt (2019) | 69.49 | 75.87 | 32.96 | 73.72 | 49.79 | 6.39 | 62.04 | 52.80 | 37.64 |
| State-of-the-art | FADA (2020) | 79.84 | 85.32 | 25.44 | 76.14 | 56.91 | 14.52 | 69.37 | 66.32 | 44.08 |
| methods | SDCA (2021) | 70.16 | 87.05 | 41.42 | 74.89 | 47.79 | 30.25 | 69.48 | 59.99 | 44.38 |
| | MUCSS (2021) | 61.33 | 83.00 | 42.17 | 70.66 | 57.88 | 13.88 | 54.82 | - | 39.93 |
| | Zhang's (2022) | 76.89 | 84.81 | **56.26** | 68.10 | 57.15 | 18.64 | 60.31 | - | 46.13 |
| | ProCA (2022) | 76.77 | **88.71** | 42.93 | 64.38 | 50.97 | 20.67 | **71.39** | 64.07 | 45.63 |
| | (Ours) | **81.06** | 84.67 | 43.34 | **81.50** | **60.08** | **39.64** | 69.76 | **71.75** | **47.67** |

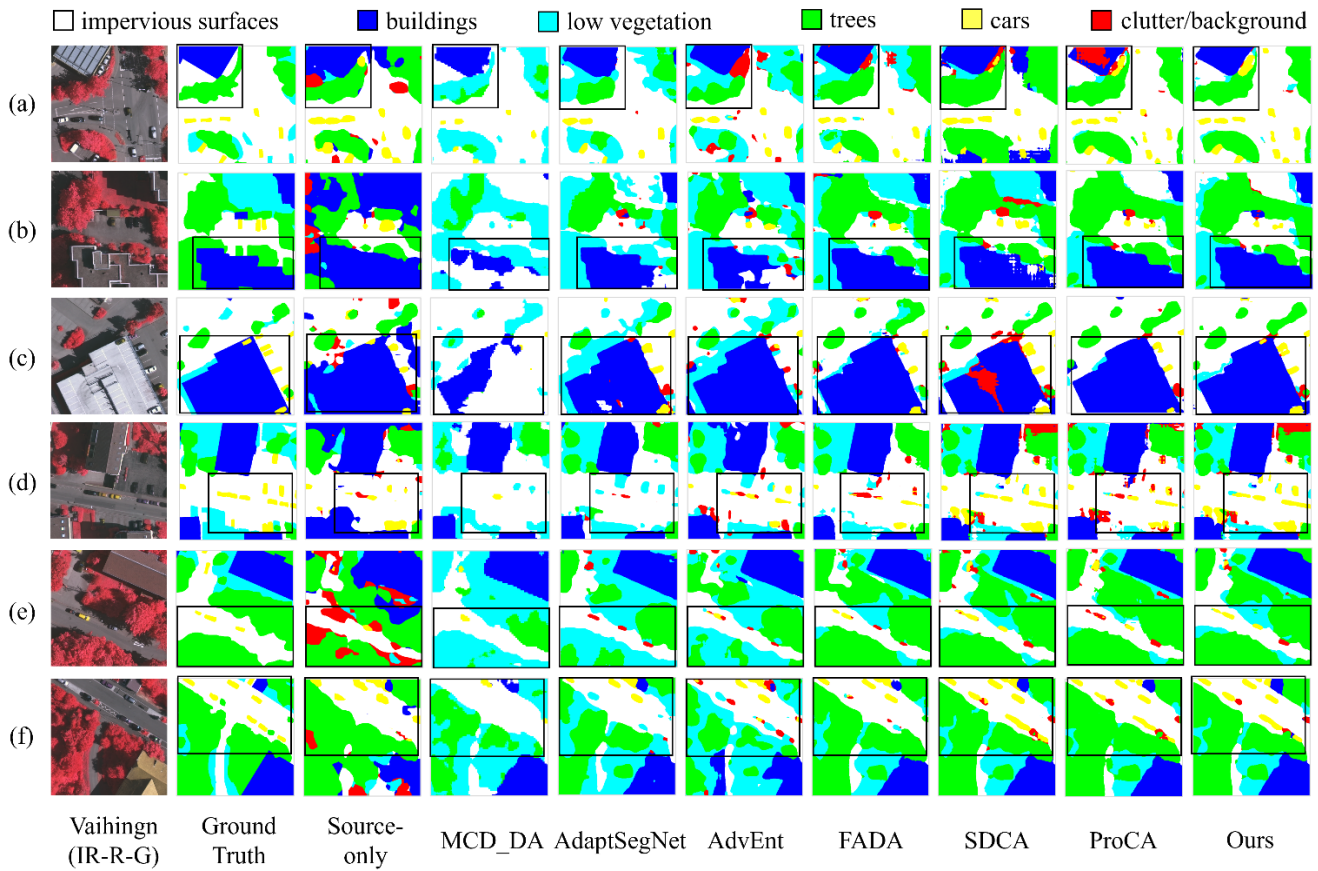

Fig. 5. Qualitative results on the POT (R-G-B) →VAI (IR-R-G) task.

Fig. 5 shows the qualitative results on the POT(IR-R-G) → VAI (IR-R-G) task. From Fig. 5, the semantic segmentation performance of Source-only model does not yet meet the requirements of practical applications because of the existence of semantic blurring and classification confusion between geo-objects. After domain adaptation, the semantic segmentation performance of all comparative state-of-the-art UDA methods is improved compared with Source-only model, but there are still problems of classification confusion and boundary blurring, for example, Building and Clutter are confused with each other in Fig. 5(a)(b)(c), and Car and Clutter are confused in Fig. 5 (d)(e)(f). However, the method proposed in this paper can better alleviate the above problems and achieve the best semantic segmentation performance. This is because the category-level causal relationship between two domain images and semantic labels can be effectively mined through the contrastive learning of category-level causal features, which can better alleviate the problem of inaccurate semantic information and confusing classification.

3) *Comparative Studies on VAI (IR-R-G) →POT (IR-R-G)*

The quantitative evaluation results of the compared domain adaptation methods on VAI (IR-R-G) →POT (IR-R-G) task are presented in Table V. From Table V, the Source-only model also has the worst performance with OA, MA, and mIoU values of 64.36%, 54.05%, and 41.22%, respectively. By considering the reduction of the domain shift between the source and target domains, eight comparative state-of-the-art UDA methods achieve better segmentation results than the Source-only model, but there is still much room for improvement. The CPCA achieves the best performance with OA, MA, and mIoU values of 73.18%, 63.01%, and 50.72%, respectively. Compared with the Source-only model, CPCA shows the performance improvement of 8.82% on OA, 8.96% on MA, and 9.50% on mIoU. Compared to the ProCA with the best performance, CPCA presents the improvement of 1.28%, 2.49%, and 2.52% on OA, MA and mIoU, respectively. This demonstrates that disentangling causal features from input images and then aligning them between the source and target domains from a causal perspective can effectively bridge the distribution shift between different domains, thus alleviating the impact of domain discrepancy on cross-domain semantic segmentation tasks.

TABLE V
QUANTITATIVE EVALUATION RESULTS (%) OF DIFFERENT UDA MODELS ON THE VAI (IR-R-G) →POT (IR-R-G) TASK

| | Methods | $F_1$ score | | | | | | OA | MA | mIoU |
|---|---|---|---|---|---|---|---|---|---|---|
| | | Imp. surf. | Building | Low veg. | Tree | Car | Clutter | | | |
| | Source-only | 71.32 | 69.45 | 60.97 | 54.67 | 69.10 | 5.47 | 64.36 | 54.05 | 41.22 |
| | MCD_DA (2018) | 75.67 | 76.72 | 63.53 | 39.97 | 60.68 | 0.00 | 67.04 | 56.42 | 43.71 |
| | AdaptSegNet (2018) | 76.55 | 77.93 | 62.15 | 45.45 | 70.89 | 6.16 | 68.93 | 57.33 | 44.94 |
| | AdvEnt (2019) | 76.65 | 82.64 | 63.73 | 32.94 | 71.74 | 2.76 | 69.59 | 58.03 | 45.43 |
| State-of-the-art | FADA (2020) | 79.04 | 82.89 | 69.31 | 29.02 | 82.75 | 5.82 | 71.20 | 59.63 | 46.75 |
| methods | SDCA (2021) | 79.35 | 82.17 | 70.10 | 41.35 | 78.76 | 5.06 | 71.74 | 60.03 | 47.18 |
| | MUCSS (2021) | 67.53 | 69.59 | 53.48 | 51.82 | 65.31 | 20.56 | 54.71 | - | 39.30 |
| | Zhang's (2022) | 78.59 | 79.84 | 63.27 | **54.60** | 75.08 | 24.59 | 62.66 | - | 47.87 |
| | ProCA (2022) | 77.81 | 81.20 | **71.94** | 37.90 | 81.78 | 15.56 | 71.90 | 60.52 | 48.20 |
| | (ours) | **81.51** | **85.59** | 71.02 | 38.00 | **84.45** | **26.68** | **73.18** | **63.01** | **50.72** |

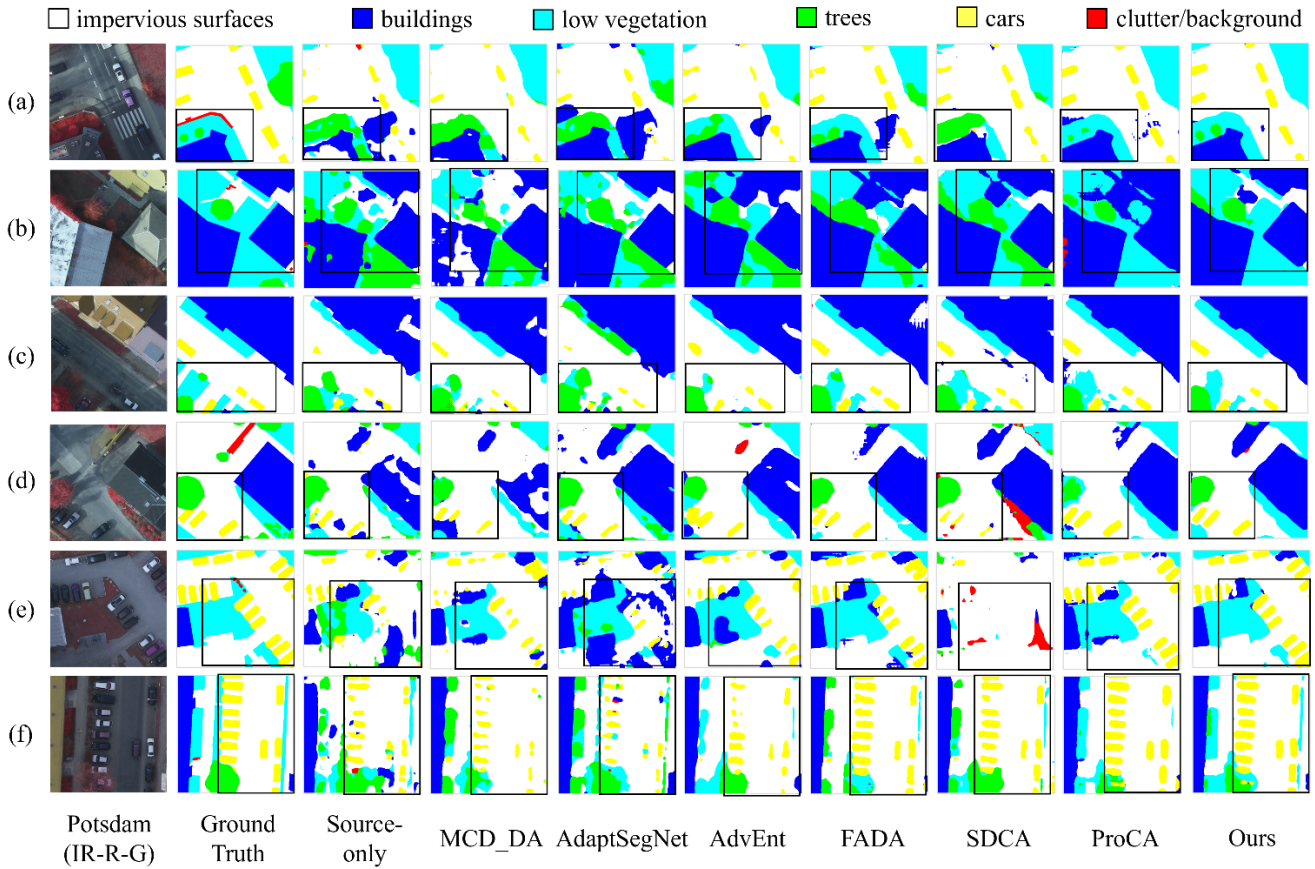

Fig. 6. Qualitative results on the VAI (IR-R-G) →POT (IR-R-G) task.

From Table V, the $F_1$ scores of CPCA in Imp. Surf., Building, Car and Clutter realize the best accuracies of 81.51%, 85.59%, 84.45% and 26.68%, respectively, and achieve the improvement of 10.19%, 16.14%, 15.35% and 21.21% compared to the Source-only model. It is worth noting that all domain adaptation methods, including the proposed one, show negative transfer to Tree. This is mainly due to the fact that the trees in the Potsdam dataset are more sparse and bare relative to the dense and tall trees in the Vaihingen dataset, resulting in objects under the trees not being shaded and presenting a more diverse visual appearance in remote sensing images. As a result, the models tend to predict the trees as other geo-object classes, such as Cars, Building, Imp. surf., etc.

We selected six typical remote sensing scenarios in the Potsdam test set for comparing the cross-domain semantic segmentation performance on the VAI (IR-R-G) →POT (IR-R-G) task, Fig. 6 shows the qualitative results. From Fig. 6, the Source-only model has poorer semantic segmentation results compared with ground truth due to the domain gap. The semantic segmentation results of the eight state-of-the-art UDA comparative methods are improved compared to Source-only model, but still suffer from confusion, boundary blurring, and cavity of geo-object in the predicted mask, such as the confusion between Imp. surf. and Building in Fig. 6(a)(b)(e), as well as between Low. veg. and Tree in Fig. 6(a)(b)(c). While our method CPCA achieves the best semantic segmentation performance, which effectively alleviate the above problems, and achieves better segmentation performance on Building, Car and Low. veg. For example, CPCA effectively solved the cavitation problem of building in (b) (c) (d), and capture the long strip features of Low. veg. more completely in (f). This shows that our CPCA can better handle the VAI (IR-R-G) → POT (IR-R-G) task.

*C. Ablation Experiments*

*1) Effectiveness of Each Component*

To evaluate the contribution of each module in the proposed CPCA, Table VI presents the ablation study results of the CFD, CPC, CI components and Self-training strategy on POT(IR-R-G) →VAI (IR-R-G) task. Compared to the Source-only model, employing the CFD module improved the results by 9.0% in OA, and 8.94% in mIoU, respectively, which confirms the validity of the disentanglement of causal features. The further integration of the CI module improved OA and mIoU to 76.86% and 53.82%. Meanwhile, compared to the Source-only model, employing the CPC improved the results by 9.65% in OA, and 9.74% in mIoU, respectively, which confirms the effectiveness of causal prototypical contrastive learning in aligning feature distribution. Notable, only employing the CPC module achieves better segmentation performance than only employing the CFD module with the improvement of 0.53% on OA, and 0.80% on mIoU, which indicates that pulling the feature distributions between the source and target domain is more effective than disentangling the causal feature from input images. Further, simultaneously employing CFD module and CPC module improved the OA and mIoU to 77.37% and 54.42%, which achieves further improvements of 1.06% on OA and 1.9% on mIoU, respectively, compared to the model only employing the CFD module, and achieves further improvements of 0.53% on OA, 1.1% on mIoU, respectively, compared to the model only employing the CPC module. The performance of OA and mIoU reaches 77.00% and 56.32%, respectively when adding CI module. Finally, we obtain the pseudo-label of the target domain based on the CPCA model and perform Self-training on the target domain on the Deeplab-v3 model with ResNet-50 and ResNet-101 backbone, respectively. It achieved further improvements of 2.26% on OA and 4.1% on mIoU with ResNet-50 backbone, and 3.18% on OA and 4.43% on mIoU with ResNet-101 backbone.

TABLE VI
ABLATION STUDY RESULTS (%) ON THE POT(IR-R-G) →VAI (IR-R-G) TASK

| Source-only | CFD | CPC | CI | Self-training | OA | mIoU |
|---|---|---|---|---|---|---|
| ✓ | | | | | 67.19 | 43.58 |
| | ✓ | | | | 76.31 | 52.52 |
| | ✓ | | ✓ | | 76.86 | 53.82 |
| | | ✓ | | | 76.84 | 53.32 |
| | ✓ | ✓ | | | 77.37 | 54.42 |
| | ✓ | ✓ | ✓ | | 77.00 | 56.32 |
| | ✓ | ✓ | ✓ | ResNet-50 ✓ | 79.26 | 60.42 |
| | ✓ | ✓ | ✓ | ResNet-101 ✓ | **80.18** | **60.75** |

*2) CAM of Each Component*

In order to visually show the importance of each component in the model, we visualize the class activation map (CAM) of features in different combinations of the model (such as CFD, CPC, CFD+CI, CFD+CPC, CFD+CPC+CI) on the POT (IR-R-G) → VAI (IR-R-G) task. From Fig 7, the model with CFD component has deviations in the feature responses of each category. For example, the response to Building involves Impervious Surfaces and Car, the response to Car is mistakenly located in Building and Tree, the response to Impervious Surfaces mistakenly involves Building, the response to low vegetation involves Impervious Surfaces, and the response to Tree is more responsive to Low Vegetation, resulting in poor semantic segmentation performance. The model with CPC component responds more accurate than the model with CFD component in the Building, Car, and Impervious Surfaces categories, but still has some response bias in the categories of

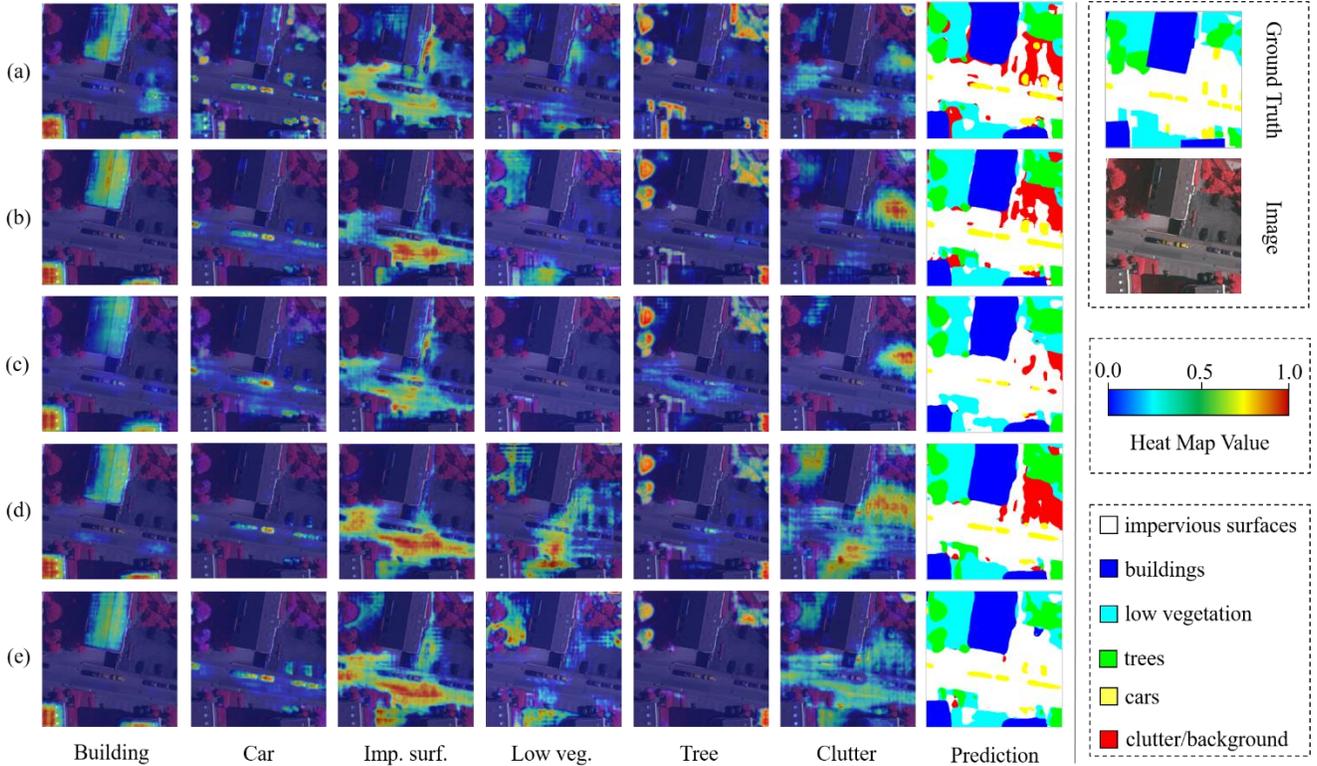

Fig. 7. CAM of Each Component on the POT (IR-R-G) →VAI (IR-R-G) task. (a) the CAM of causal encoder of the model with CFD components. (b) the CAM of the model with CPC components. (c) the CAM of causal encoder of the model with CFD+CI components. (d) the CAM of causal encoder of the model with CFD+CPC components. (e) the CAM of causal encoder of the model with CFD+CPC+CI components.

TABLE VII
ABLATION STUDIES OF DIFFERENT CAUSAL FEATURE ALIGNMENT METHODS FOR POT (IR-R-G) →VAI (IR-R-G) TASK.

| Methods | $F_1$ score | | | | | | OA | MA | mIoU |
|---|---|---|---|---|---|---|---|---|---|
| | Imp. surf. | Building | Low veg. | Tree | Car | Clutter | | | |
| Source-only | 72.85 | 74.40 | 56.42 | 67.96 | 46.76 | 28.49 | 67.19 | 60.55 | 43.58 |
| GAT (global-wise adversarial training) | **81.96** | 88.45 | 53.27 | 76.15 | 51.89 | 51.72 | 75.79 | 68.10 | 52.74 |
| FADA (class-wise adversarial training) | 81.18 | **89.03** | 57.80 | 77.86 | 56.65 | 44.04 | 76.82 | 69.90 | 53.45 |
| Ours (contrast training) | 78.40 | 88.32 | **62.23** | **78.66** | **59.36** | **59.33** | **77.00** | **71.31** | **56.32** |

Impervious Surfaces and Tree. Further, the models with CFD+CI component and CFD+CPC component can achieve better feature response and semantic segmentation performance than (a) and (b). Finally, by combining the CFD, CPC, and CI components, the model achieves the most accurate feature response in each category, which leads to the best semantic segmentation results.

*3) Effectiveness of Causal Prototypical Contrast Module*

To verify the effectiveness of the CPC module, we implement other causal feature alignment methods, such as global-wise adversarial training GAT and class-wise adversarial training FADA. From Table VII, GAT improves the baseline to 75.79% on OA, 68.10% on MA, and 52.74% on mIoU, respectively, which indicates the effectiveness of the global-wise adversarial training in narrowing domain shift. FADA obtains the performance with OA, MA, and mIoU values of 76.82%, 69.90%, and 53.45%, respectively, by considering class-wise feature alignment. Compared with above methods, CPCA achieves the best performance with OA, MA, and mIoU values of 77.00%, 71.31%, and 56.32%, respectively, which demonstrates the superiority of the class-wise contrastive training than global-wise and class-wise causal adversarial training.

*D. Visualization of Causal Features and Bias Features*

We conduct visualization experiments on the causal and bias features of CPCA, and the results are shown in Fig. 8, where there are some differences and complementarities between the causal and bias features of CPCA. For example, the building causal features in Fig. 8(a)(b) pay more attention to the global information of the building, and respond more uniformly to the features in the building area, while the bias features are more focused on the information of the local details and respond more to the information of the surrounding geo-objects. The causal and bias features of Car are more complementary in scenario (a), with the causal features corresponding to the car and the bias features focusing more on the tree next to the car. In scenario (b), the causal features respond more accurately to car and also consider the co-occurrence relationship between Car and Tree, Imp. Surf. For the categories of Imp. Surf., Low veg. and tree, their causal features respond themselves more comprehensively and accurately, whereas the bias features do not accurately capture the location of each category. As for the Clutter, causal features can accurately localize its region in scenario (a), while bias features focus more on other categories. Since there is no clutter category in scenario (b), both causal and bias features respond to information of other categories, but the response of causal features is slightly lower. In summary, the causal and bias features of CPCA focus on different information of geo-objects, in which the causal features pay more attention to information of geo-objects themselves, with accurate and comprehensive responses; while the bias features focus on the local and chaotic information of geo-objects with less precise and scattered responses.

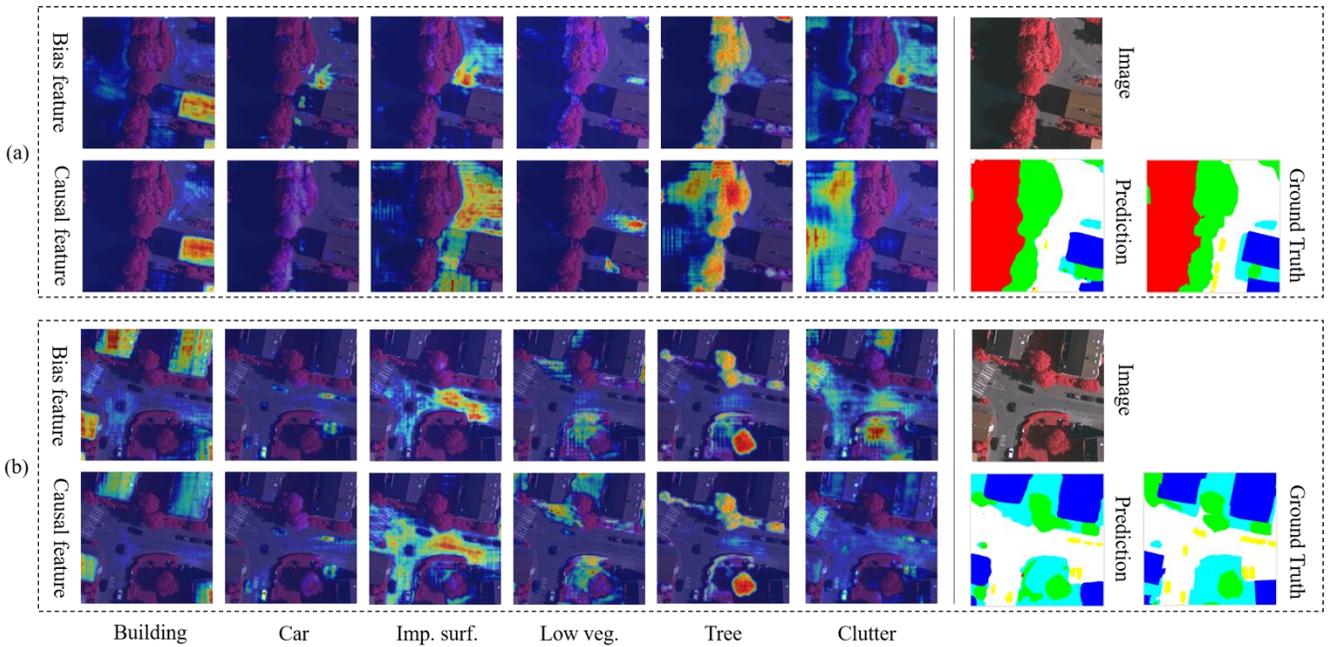

Fig. 8. Visualization of causal feature and non-causal feature on the POT (IR-R-G) →VAI (IR-R-G) task.

## V. Conclusion And Future Work

In this paper, we meet the UDA semantic segmentation challenge from a causal view, and propose a novel causal prototype-inspired contrast adaptation (CPCA) method. CPCA assumes that each source- and target-domain remote sensing image is composed of causal and non-causal factors. Since it is difficult to precisely separate causal and non-causal factors from the original input, we attempt to disentangle the domain-invariant features from the source and target domain images by disentangled representations learning, and force the domain-invariant features to replace causal factors by satisfying the following three principles: 1) causal factors should be separated from the non-causal factors; 2) the causal factors of source and target domains are domain invariant; 3) intervening on the non-causal factors will not affect the causal factors to make decisions on labels. Through the above attempts, the CPCA can build invariant causal mechanism between source and target domains, which enhances the generalization of model on target domain. Comprehensive experiments demonstrate the validity, interpretability, and generalizability of the proposed method.

Since not all causal information is most critical in domain adaptation tasks, in the future, we aim to further explore the sufficient and necessary causes rather than all possible causes.